\documentclass[letterpaper]{article}
\usepackage{aaai18}

\usepackage[hidelinks]{hyperref}

\usepackage{natbib}
\newcommand{\newcite}{\citet}
\renewcommand{\cite}{\citep}

\usepackage{times}
\usepackage{helvet}
\usepackage{courier}
\usepackage{color}
\usepackage{url}
\usepackage{hyperref}
\usepackage{danudefs}
\usepackage{graphicx}
\usepackage{caption}
\usepackage{subcaption}
\usepackage{booktabs}

\usepackage{todonotes}

\usepackage[boldmath]{numprint}
\usepackage{multirow}
\usepackage{numprint}
\npdecimalsign{.}
\nprounddigits{2}
\newcolumntype{H}{>{\setbox0=\hbox\bgroup}c<{\egroup}@{}}

\usepackage{colortbl}
\definecolor{LightCyan}{rgb}{0.88,1,1}

\usepackage{algorithmic}
\usepackage{algorithm}

\begin{document}
%
\title{Think Globally, Embed Locally --- Locally Linear Meta-embedding of Words}

\author{Danushka Bollegala\footnotemark[1]\footnotemark[4] \ \ 
Kohei Hayashi\footnotemark[2]\footnotemark[4] \ \ Ken-ichi Kawarabayashi\footnotemark[3]\footnotemark[4]\\
The University of Liverpool, Liverpool, United Kingdom.\footnotemark[1]  \\
Japan Advanced Institute of Science and Technology, Tokyo, Japan.\footnotemark[2]\\
National Institute of Informatics, Tokyo, Japan. \footnotemark[3]\\
Japan Science and Technology Agency, ERATO, Kawarabayashi Large Graph Project.\footnotemark[4]
}


\maketitle
\begin{abstract}
Distributed word embeddings have shown superior performances in numerous Natural Language Processing (NLP) tasks.
However, their performances vary significantly across different tasks,
implying that the word embeddings learnt by those methods capture complementary aspects of lexical semantics.
Therefore, we believe that it is important to combine the existing word embeddings to produce more accurate and complete
\emph{meta-embeddings} of words.
For this purpose, we propose an unsupervised locally linear meta-embedding learning method that takes 
pre-trained word embeddings as the input, and produces more accurate meta embeddings.
Unlike previously proposed meta-embedding learning methods that learn a global projection over all words in a vocabulary,
our proposed method is sensitive to the differences in local neighbourhoods of the individual source word embeddings. 
Moreover, we show that vector concatenation, a previously proposed highly competitive baseline
approach for integrating word embeddings, can be derived as a special case of the proposed method.
Experimental results on semantic similarity, word analogy, relation classification, and short-text classification tasks show that our
meta-embeddings to significantly outperform prior methods in several benchmark datasets,
establishing a new state of the art for meta-embeddings.
\end{abstract}

\section{Introduction}
\label{sec:intro}

Representing the meanings of words is a fundamental task in Natural Language Processing (NLP). 
One popular approach to represent the meaning of a word is to \emph{embed} it in some fixed-dimensional vector space~\cite{Turney:JAIR:2010}. 
In contrast to sparse and high-dimensional counting-based distributional word representation methods  that use co-occurring contexts of a word
as its representation~\cite{baroni-dinu-kruszewski:2014:P14-1},  dense and low-dimensional prediction-based distributed word representations have obtained impressive performances in numerous NLP tasks such as sentiment classification~\citep{socher-EtAl:2013:EMNLP}, and machine translation~\cite{Zou:EMNLP:2013}.
Several distributed word embedding learning methods based on different learning strategies have
been proposed~\cite{Pennington:EMNLP:2014,Milkov:2013,Huang:ACL:2012,Collobert:ICML:2008,Mnih:HLBL:NIPS:2008}.

Previous works studying the differences in word embedding learning methods~\citep{Chen:2013,Yin:ACL:2016} have shown that word embeddings learnt using different methods and from different resources have significant variation in quality and characteristics of the semantics captured.
For example, \newcite{Hill:NIPS:2014,Hill:ICLR:2015} showed that the word embeddings trained from monolingual vs. bilingual corpora 
capture different local neighbourhoods. \newcite{Bansal:ACL:2014} showed that an ensemble of different word representations
improves the accuracy of dependency parsing, implying the complementarity of the different word embeddings.
This suggests the importance of \emph{meta-embedding} -- creating a new embedding by combining different existing embeddings.
We refer to the input word embeddings to the meta-embedding process as the \emph{source embeddings}.
\newcite{Yin:ACL:2016} showed that by meta-embedding five different pre-trained word embeddings,
 we can overcome the out-of-vocabulary problem, and improve the accuracy of cross-domain part-of-speech (POS) tagging.
Encouraged by the above-mentioned prior results, we expect an ensemble containing multiple word embeddings to produce
better performances than the constituent individual embeddings in NLP tasks.

There are three main challenges a meta-embedding learning method must overcome.

First, the vocabularies covered by the source embeddings might be different because they have been trained on different text corpora. 
Therefore, not all words will be equally represented by all the source embeddings.
Even in situations where the implementations of the word embedding learning methods are publicly available,
it might not be possible to retrain those embeddings because the text corpora on which those methods were originally trained might not be publicly available.
Moreover, it is desirable if the meta-embedding method does not require the original resources upon which they were trained such as
corpora or lexicons, and can directly work with the pre-trained word embeddings. 
This is particularly attractive from a computational point of view because re-training source embedding methods on large corpora might require significant processing times and resources.

Second, the vector spaces and their dimensionalities of the source embeddings might be different.
In most prediction-based word embedding learning methods the word vectors are randomly initialised.
Therefore, there is no obvious correspondence between the dimensions in two word embeddings learnt even from two different runs of the same method, let alone from different methods~\citep{Tian:arxiv:2016}.
Moreover, the pre-trained word embeddings might have different dimensionalities, which is often a hyperparameter set experimentally.
This becomes a challenging task when incorporating multiple source embeddings to learn a single meta-embedding because the alignment between the dimensionalities of the source embeddings is unknown.

Third, the local neighbourhoods of a particular word under different word embeddings show a significant diversity.
For example, as the nearest neighbours of the word \emph{bank}, 
GloVe~\citep{Pennington:EMNLP:2014}, a word sense insensitive embedding, lists \emph{credit, financial, cash},
whereas word sense sensitive embeddings created by \newcite{Huang:ACL:2012} lists \emph{river, valley, marsh}
when trained on the same corpus. 
We see that the nearest neighbours for the different senses of the word bank (i.e. financial institution vs. river bank) are
captured by the different word embeddings. 
Meta-embedding learning methods that learn a single global projection over the entire vocabulary are insensitive to such local variations in the neighbourhoods~\citep{Yin:ACL:2016}.

To overcome the above-mentioned challenges, we propose a \emph{locally-linear} meta-embedding learning method that
(a) requires only the words in the vocabulary of each source embedding, without having to predict embeddings for missing words,
(b) can meta-embed source embeddings with different dimensionalities,
(c) is sensitive to the diversity of the neighbourhoods of the source embeddings.

Our proposed method comprises of two steps: a neighbourhood \emph{reconstruction step} (Section~\ref{sec:nn}), and a \emph{projection step} (Section~\ref{sec:proj}).
In the reconstruction step, we represent the embedding of a word by the linearly weighted combination of the embeddings of its nearest neighbours in each source embedding space.
Although the number of words in the vocabulary of a particular source embedding can be potentially large, the consideration of nearest neighbours
enables us to limit the representation to a handful of parameters per each word, not exceeding the neighbourhood size.
The weights we learn are shared across different source embeddings, thereby incorporating the information from different source
embeddings in the meta-embedding. 
Interestingly, vector concatenation, which has found to be an accurate meta-embedding method, can be derived as a special case of
this reconstruction step.

Next, the projection step computes the meta-embedding of each word such that the nearest neighbours in the source embedding spaces are embedded closely to each other in the meta-embedding space.
The reconstruction weights can be efficiently computed using stochastic gradient descent, whereas the projection can be efficiently
computed using a truncated eigensolver. 

It is noteworthy that we do not directly compare different source embeddings for the same word in the reconstruction step nor in the projection step. 
This is important because the dimensions in source word embeddings learnt using different word embedding learning methods are not aligned. 
Moreover, a particular word might not be represented by all source embeddings. 
This property of the proposed method is attractive because it obviates the need to align source embeddings, or predict missing source word embeddings prior to meta-embedding. Therefore, all three challenges described above are solved by the proposed method.

The above-mentioned properties of the proposed method enables us to compute meta-embeddings 
for five different source embeddings covering 2.7 million unique words.
We evaluate the meta-embeddings learnt by the proposed method on 
semantic similarity prediction, analogy detection, relation classification, and short-text classification tasks. 
The proposed method significantly outperforms several competitive baselines and 
previously proposed meta-embedding learning methods~\citep{Yin:ACL:2016} on multiple benchmark datasets. 

\section{Related Work}
\label{sec:related}

\newcite{Yin:ACL:2016} proposed a meta-embedding learning method (1\texttt{TO}N) that projects a meta-embedding of a word into the source embeddings using separate projection matrices. 
The projection matrices are learnt by minimising the sum of squared Euclidean distance between the projected source embeddings and the corresponding original source embeddings for all the words in the vocabulary. 
They propose an extension (1\texttt{TO}N+) to their meta-embedding learning method that first predicts the source word embeddings for out-of-vocabulary words in a particular source embedding, using the known word embeddings. 
Next, 1\texttt{TO}N method is applied to learn the meta-embeddings for the union of the vocabularies covered by all of the source embeddings.

Experimental results in semantic similarity prediction, word analogy detection, and cross-domain POS tagging tasks show the effectiveness of both 1\texttt{TO}N and 1\texttt{TO}N+. In contrast to our proposed method which learns locally-linear projections that are sensitive to the variations in the local neighbourhoods in the source embeddings,
1\texttt{TO}N and 1\texttt{TO}N+ can be seen as globally linear projections between meta and source embedding spaces. 
As we see later in Section~\ref{sec:res}, our proposed method outperforms both of those methods consistently in all benchmark tasks
demonstrating the importance of neighbourhood information when learning meta-embeddings. 
Moreover, our proposed meta-embedding method does not directly compare different source embeddings, thereby obviating the need to predict source embeddings for out-of-vocabulary words.
Locally-linear embeddings are attractive from a computational point-of-view as well because during optimisation
we require information from only the local neighbourhood of each word.

Although not learning any meta-embeddings, several prior work have shown that by incorporating multiple word embeddings learnt using different methods improve performance in various NLP tasks. 
For example, \newcite{tsuboi:2014:EMNLP2014} showed that by using both word2vec and GloVe embeddings together in a POS tagging task, it is possible to improve the tagging accuracy, if we had used only one of those embeddings. 
Similarly, \newcite{Turian:ACL:2010} collectively used Brown clusters, CW and HLBL embeddings, to improve the performance of named entity recognition and chucking tasks.

\newcite{Luo:AAAI:2014} proposed a multi-view word embedding learning method that uses a two-sided neural network.
They adapt pre-trained CBOW~ \citep{Mikolov:NIPS:2013}  embeddings from Wikipedia and click-through data from a search engine.
Their problem setting is different from ours because their source embeddings are trained using the same word embedding learning method but on different resources whereas, we consider source embeddings trained using different word embedding learning methods and resources.
Although their method could be potentially extended to meta-embed different source embeddings, the unavailability of their implementation
prevented us from exploring this possibility.

\newcite{AAAI:2016:Goikoetxea} showed that concatenation of word embeddings learnt separately from a corpus and the WordNet
to produce superior word embeddings. Moreover, performing Principal Component Analysis (PCA) on the concatenated embeddings
slightly improved the performance on word similarity tasks. In Section~\ref{sec:baselines}, we discuss the relationship between the proposed
method and vector concatenation.

\section{Locally Linear Meta-Embeddings}
\label{sec:method}

\subsection{Problem Settings}
\label{sec:definition}

To explain the proposed meta-embedding learning method, let us consider two source word embeddings, denoted by $\cS_{1}$ and $\cS_{2}$.
Although we limit our discussion here to two source embeddings for the simplicity of the description,
the proposed meta-embedding learning method can be applied to any number of source embeddings.
Indeed in our experiments we consider five different source embeddings.
Moreover, the proposed method is not limited to meta-embedding unigrams, and can be used for $n$-grams of any length $n$,
provided that we have source embeddings for those $n$-grams.

We denote the dimensionalities of $\cS_{1}$ and $\cS_{2}$ respectively by $d_{1}$ and $d_{2}$ (in general, $d_{1} \neq d_{2}$). 
The sets of words covered by each source embedding (i.e. vocabulary) are denoted by $\cV_{1}$ and $\cV_{2}$.
The source embedding of a word $v \in \cV_{1}$ is represented by a vector $\vec{v}^{(1)} \in \R^{d_{1}}$,
whereas the same for a word $v \in \cV_{2}$ by a vector $\vec{v}^{(2)} \in \R^{d_{2}}$.
Let the set union of $\cV_{1}$ and $\cV_{2}$ be $\cV = \{ v_{1}, \ldots, v_{n}\}$ containing $n$ words.
In particular, note that our proposed method does not require a word $v$ to be represented by all source embeddings, and can operate on the union of the vocabularies of the source embeddings.
The meta-embedding learning problem is then to learn an embedding $\vec{v}^{(\cP)} \in \R^{d_{\cP}}$ in a meta-embedding space $\cP$ with dimensionality $d_{\cP}$ for each word $v \in \cV$.

For a word $v$, we denote its $k$-nearest neighbour set in embedding spaces $\cS_{1}$ and $\cS_{2}$ respectively by $\cN_{1}(v)$ and $\cN_{2}(v)$
(in general, $|\cN_{1}(v)| \neq |\cN_{2}(v)|$).
As discussed already in Section~\ref{sec:intro}, different word embedding methods encode different aspects of lexical semantics, and are likely to have different local neighbourhoods.
Therefore, by requiring the meta embedding to consider different neighbourhood constraints in the source embedding spaces we hope to
exploit the complementarity in the source embeddings.

\subsection{Nearest Neighbour Reconstruction}
\label{sec:nn}

The first-step in learning a locally linear meta-embedding is to reconstruct each source word embedding using a linearly weighted combination
 of its $k$-nearest neighbours. 
Specifically, we construct each word $v \in \cV$ separately from its $k$-nearest neighbours $\cN_{1}(v)$, and $\cN_{2}(v)$. 
The reconstruction weight $w_{vu}$  assigned to a neighbour $u \in \cN_{1}(v) \cup \cN_{2}(v)$ is found by minimising the reconstruction error $\Phi(\mat{W})$ defined by \eqref{eq:weights}, which is the sum of local distortions in the two source embedding spaces.
\begin{align}
\label{eq:weights}
\Phi(\mat{W}) =  \sum_{i=1}^{2}\sum_{v \in \cV} \norm{\vec{v}^{(i)} - \sum_{u \in \cN_{i}(v)} w_{vu} \vec{u}^{(i)}}_{2}^{2} 
\end{align}
Words that are not $k$-nearest neighbours of $v$ in either of the source embedding spaces will have their weights set to zero 
(i.e. $w_{vu} = 0, \forall u \notin \cN_{1}(v) \cup \cN_{2}(v)$).
Moreover, we require the sum of reconstruction weights for each $v$ to be equal to one 
(i.e. $\sum_{u \in \cV} w_{uv} = 1$). 

To compute the weights $w_{vu}$ that minimise \eqref{eq:weights}, we compute its error gradient $\frac{\partial \Phi(\mat{W})}{\partial w_{vu}}$ as follows:
\begin{align*}
\label{eq:dW}
-2\sum_{i=1}^{2}{\left(\vec{v}^{(i)} - \sum_{x \in \cN_{i}(v)} w_{vx} \vec{x}^{(i)}\right)}^{\top}{\vec{u}^{(i)}}\mathbb{I}[u \in \cN_{i}(v)] 
\end{align*}
Here, the indicator function, $\mathbb{I}[x]$, returns $1$ if $x$ is true and $0$ otherwise.
We uniformly randomly initialise the weights $w_{vu}$ for each neighbour $u$ of $v$, and use stochastic gradient descent (SGD)
with the learning rate scheduled by AdaGrad~\citep{Duchi:JMLR:2011} to compute the optimal values of the weights. 
The initial learning rate is set to $0.01$ and the maximum number of iterations to $100$ in our experiments. 
Empirically we found that these settings to be adequate for convergence. 
Finally, we normalise the weights $w_{uv}$ for each $v$ such that they sum to 1 (i.e. $\sum_{u \in \cV} w_{vu} = 1$). 

Exact computation of $k$ nearest neighbours for a given data point in a set of $n$ points requires all pairwise similarity computations.
Because we must repeat this process for each data point in the set, this operation would require a time complexity of $\O(n^{3})$.
This is prohibitively large for the vocabularies we consider in NLP where typically $n>10^{3}$.
Therefore, we resort to approximate methods for computing $k$ nearest neighbours.
Specifically, we use the BallTree algorithm~\citep{BallTree} to efficiently compute the approximate $k$-nearest neighbours, for which 
the time complexity of tree construction is $\O(n \log n)$ for $n$ data points.

The solution to the least square problem given by \eqref{eq:weights} subjected to the summation constraints can be found by solving
a set of linear equations. Time complexity of this step is $\cO(N (d_{1} |\cN_{1}|^{3} + d_{2} |\cN_{2}|^{3}))$,
which is cubic in the neighbourhood size and linear in both the dimensionalities of the embeddings and vocabulary size.
However, we found that the iterative estimation process using SGD described above to be more efficient in practice.
Because $k$ is significantly smaller than the number of words in the vocabulary, 
and often the word being reconstructed is contained in the neighbourhood,
the reconstruction weight computation converges after a small number (less than 5 in our experiments) of iterations.

\subsection{Projection to Meta-Embedding Space}
\label{sec:proj}

In the second step of the proposed method, we compute the meta-embeddings $\vec{v}^{(\cP)}, \vec{u}^{(\cP)} \in \R^{d_{\cP}}$ for words $v, u \in \cV$ using the reconstruction weights $w_{vu}$ we computed in Section~\ref{sec:nn}.
Specifically, the meta-embeddings must minimise the projection cost, $\Psi(\cP)$, defined by \eqref{eq:meta}.
\begin{equation}
\label{eq:meta}
\Psi(\cP) = \sum_{v \in \cV} \norm{\vec{v}^{(\cP)} - \sum_{i=1}^{2}\sum_{u \in \cN_{i}(v)} w_{vu}\vec{u}^{(\cP)}}_{2}^{2}
\end{equation}
By finding a $\cP$ space that minimises \eqref{eq:meta}, we hope to preserve the rich neighbourhood diversity in all source embeddings 
within the meta-embedding.
The two summations in \eqref{eq:meta} over $N_{1}(v)$ and $N_{2}(v)$ can be combined to re-write
\eqref{eq:meta} as follows:
\begin{equation}
\label{eq:meta2}
\Psi(\cP) = \sum_{v \in \cV} \norm{\vec{v}^{(\cP)} - \sum_{u \in \cN_{1}(v) \cup \cN_{2}(v)} w'_{vu} \vec{u}^{(\cP)}}_{2}^{2}
\end{equation}
Here, $w'_{uv}$ is computed using \eqref{eq:wprime}.
\begin{equation}
\label{eq:wprime}
w'_{vu} = w_{vu}\sum_{i=1}^{2} \mathbb{I}[u \in \cN_{i}(v)]  
\end{equation}

The $d_{\cP}$ dimensional meta-embeddings are given by the eigenvectors corresponding to the smallest $(d_{\cP} + 1)$ eigenvectors of
the matrix $\mat{M}$ given by \eqref{eq:matM}.
\begin{equation}
 \label{eq:matM}
 \mat{M} = (\mat{I} - \mat{W}')\T(\mat{I} - \mat{W}')
\end{equation}
Here, $\mat{W}'$ is a matrix with the $(v,u)$ element set to $w'_{vu}$. 
The smallest eigenvalue of $\mat{M}$ is zero and the corresponding eigenvector is discarded from the projection.
The eigenvectors corresponding to the next smallest $d_{\cP}$ eigenvalues of the symmetric matrix $\mat{M}$ can be found without performing a full matrix diagonalisation~\citep{Bai:2000}.  Operations involving $\mat{M}$ such as the left multiplication by $\mat{M}$, which is required by most sparse eigensolvers, can exploit the fact that $\mat{M}$ is expressed in \eqref{eq:matM} as the product between two sparse matrices.
Moreover, truncated randomised methods~\cite{Halko:2010} can be used to find the smallest eigenvectors, without performing full eigen decompositions.
In our experiments, we set the neighbourhood sizes for all words in all
source embeddings equal to $n$ (i.e $\forall i \  |\cN_{i}(v)| = N, \forall v \in \cV$), 
and project to a $d_{\cP} (< \! N)$ dimensional meta-embedding space.

\section{Experiments and Results}

\subsection{Source Word Embeddings}

We use five previously proposed pre-trained word embedding sets as the source embeddings in our experiments:
\begin{description}
\item (a) \textbf{HLBL} -- hierarchical log-bilinear~\citep{Mnih:HLBL:NIPS:2008} embeddings released by \newcite{Turian:ACL:2010}
 (246,122 word embeddings, 100 dimensions, trained on Reuters Newswire (RCV1) corpus),
\item (b) \textbf{Huang} -- \newcite{Huang:ACL:2012} used global contexts to train multi-prototype word embeddings that are sensitive to word senses
(100,232 word embeddings, 50 dimensions, trained on April 2010 snapshot of Wikipedia),
\item (c) \textbf{GloVe} -- \newcite{Pennington:EMNLP:2014} used global co-occurrences of words over a corpus to learn word embeddings
(1,193,514 word embeddings, 300 dimensions, trained on 42 billion corpus of web crawled texts),
\item (d) \textbf{CW} -- \newcite{Collobert:ICML:2008} learnt word embeddings following a multitask learning approach covering multiple NLP tasks
(we used the version released by  \cite{Turian:ACL:2010} trained on the same corpus as \textbf{HLBL} containing 268,810 word embeddings, 200 dimensions),
\item (e) \textbf{CBOW} -- \newcite{Mikolov:NIPS:2013} proposed the continuous bag-of-words method to train word embeddings (we discarded phrase embeddings and selected 929,922 word embeddings, 300 dimensions, trained on the Google News corpus containing ca. 100 billion words).
\end{description}
The intersection of the five vocabularies is 35,965 words, whereas their union is 2,788,636.
Although any word embedding can be used as a source we select the above-mentioned word embeddings because
(a) our goal in this paper is \emph{not} to compare the differences in performance of the source embeddings, 
and (b) by using the same source embeddings as in prior work~\cite{Yin:ACL:2016}, we can perform a fair evaluation.\footnote{Although skip-gram embeddings are shown to outperform most other embeddings, they were not used as a source by \newcite{Yin:ACL:2016}. Therefore, to be consistent in comparisons against prior work, we decided not to include skip-gram as a source.}
In particular, we could use word embeddings trained by the same algorithm but on different resources, or different algorithms on the same resources as the source embeddings. We defer such evaluations to an extended version of this conference submission.

\subsection{Evaluation Tasks}
\label{sec:eval-tasks}

The standard protocol for evaluating word embeddings is to use the embeddings in some NLP task
and to measure the relative increase (or decrease) in performance in that task.
We use four such extrinsic evaluation tasks:
\begin{description}
\item[Semantic similarity measurement:] We measure the similarity between two words as the cosine similarity between the corresponding
embeddings, and measure the Spearman correlation coefficient against the human similarity ratings. We use
Rubenstein and Goodenough's dataset~\cite{RG} (\textbf{RG}, 65 word-pairs),
rare words dataset (\textbf{RW}, 2034 word-pairs)~\citep{Luong:CoNLL:2013},
Stanford's contextual word similarities (\textbf{SCWS}, 2023 word-pairs)~\citep{Huang:ACL:2012},  
the \textbf{MEN} dataset (3000 word-pairs)~\citep{MEN}, and the SimLex dataset \citep{SimLex} (\textbf{SL} 999 word-pairs).

In addition, we use the Miller and Charles' dataset~\cite{MC} (\textbf{MC}, 30 word-pairs) as a validation dataset to tune various hyperparameters
such as the neighbourhood size, and the dimensionality of the meta-embeddings for the proposed method and baselines.

\item[Word analogy detection:] Using the CosAdd method, we solve word-analogy questions in
the Google dataset (\textbf{GL})~\citep{Mikolov:NIPS:2013} (19544 questions), and in the SemEval (\textbf{SE}) dataset~\citep{SemEavl2012:Task2}.
Specifically, for three given words $a$, $b$ and $c$, we find a fourth word $d$ that correctly answers the question
\emph{$a$ to $b$ is $c$ to what?} such that the cosine similarity between the two vectors $(\vec{b} - \vec{a} + \vec{c})$ and $\vec{d}$
is maximised. 

\item[Relation classification:] We use the DiffVec (\textbf{DV})~\citep{Vylomova:ACL:2016} dataset containing 12,458 triples 
of the form $(\textrm{relation}, \textrm{word}_{1}, \textrm{word}_{2})$ covering 15 relation types. We train a 1-nearest neighbour classifer
where for each target tuple we measure the cosine similarity between the vector offset for its two word embeddings, and
those of the remaining tuples in the dataset.
If the top ranked tuple has the same relation as the target tuple, then it is considered to be a correct match.
We compute the (micro-averaged) classification accuracy over the entire dataset as the evaluation measure.

\item[Short-text classification:]  We use two binary short-text classification datasets: Stanford sentiment treebank (\textbf{TR})\footnote{\url{http://nlp.stanford.edu/sentiment/treebank.html}} (903 positive test instances and 903 negative test instances), and the
movie reviews dataset (\textbf{MR})~\citep{Pang:ACL:2005} (5331 positive instances and 5331 negative instances).
Each review is represented as a bag-of-words and we compute the centroid of the embeddings of the words in each bag to represent that review.
Next, we train a binary logistic regression classifier with a cross-validated $\ell_{2}$ regulariser using the train portion of each dataset, and evaluate the classification accuracy using the test portion of the dataset.

\end{description}

\subsection{Baselines}
\label{sec:baselines}

\paragraph{Concatenation (CONC):}

A simple baseline method for combining pre-trained word embeddings is to concatenate the embedding vectors for a word $w$ to produce a meta-embedding for $w$. Each source embedding of $w$ is $\ell_{2}$ normalised prior to concatenation such that each source embedding contributes
equally (a value in $[-1,1]$) when measuring the word similarity using the dot product. 
As also observed by \newcite{Yin:ACL:2016}
we found that \textbf{CONC} performs poorly without emphasising \textbf{GloVe} and \textbf{CBOW} by a constant factor
(which is set to $8$ using \textbf{MC} as a validation dataset)
 when used in conjunction with \textbf{HLBL}, \textbf{Huang}, and \textbf{CW} source embeddings.
 
Interestingly, concatenation can be seen as a special case in the reconstruction step described in Section~\ref{sec:nn}.
To see this, let us denote the concatenation of column vectors 
$\vec{v}^{(1)}$ and $\vec{v}^{(2)}$ by $\vec{x} = (\vec{v}^{(1)}; \vec{v}^{(2)})$,
and $\vec{u}^{(1)}$ and $\vec{u}^{(2)}$ by $\vec{y} = (\vec{u}^{(1)}; \vec{u}^{(2)})$, where $\vec{x}, \vec{y} \in \R^{d_{1} + d_{2}}$.
Then, the reconstruction error defined by \eqref{eq:weights} can be written as follows:
\begin{equation}
\label{eq:conc}
\Phi(\mat{W}) = \sum_{v \in \cV} \norm{\vec{x} - \sum_{u \in \cN(v)} w_{vu}\vec{y} }_{2}^{2}
\end{equation}
Here, the vocabulary $\cV$ is constrained to the intersection $\cV_{\cS} \cap \cV_{\cT}$ because concatenation is not defined for
missing words in a source embedding. Alternatively, one could use zero-vectors for missing words or (better) predict the word embeddings
for missing words prior to concatenation. However, we consider such extensions to be beyond the simple concatenation baseline we consider here.\footnote{Missing words does \emph{not} affect the performance of \textbf{CONC} because all words in the benchmark datasets we use in our experiments are covered by all source embeddings.}
On the other hand, the common neighbourhood $\cN(v)$ in \eqref{eq:conc} can be obtained by either
 limiting $\cN(v)$ to $\cN_{1}(v) \cap \cN_{2}(v)$ or, by extending the neighbourhoods to the entire vocabulary ($\cN(v) = \cV$).
  \eqref{eq:conc} shows that under those neighbourhood constraints, the first step in our proposed method can be seen as reconstructing
 the neighbourhood of the concatenated space.  The second step would then find meta-embeddings that preserve the locally linear structure in the
 concatenated space.
 
One drawback of concatenation is that it increases the dimensionality of the meta-embeddings compared to the source-embeddings,
which might be problematic when storing or processing the meta-embeddings 
(for example, for the five source embeddings we use here $d_{\cP} = 100 + 50 + 300 + 200 + 300 = 950$).

\paragraph{Singular Value Decomposition (SVD):}

We create an $N \times 950$ matrix $\mat{C}$ by arranging the \textbf{CONC} vectors for the union of all source embedding vocabularies.
For words that are missing in a particular source embedding, we assign zero vectors of that source embedding's dimensionality.
Next, we perform SVD on $\mat{C} = \mat{U} \mat{D} \mat{V}\T$, where $\mat{U}$ and $\mat{V}$ are unitary matrices and
the diagonal matrix $\mat{D}$ contains the singular values of $\mat{C}$. We then select the $d$ largest left singular vectors from 
$\mat{U}$ to create a $d_{\cP}$ dimensional embeddings for the $N$ words. Using the \textbf{MC} validation dataset, we set $d_{\cP} = 300$.
Multiplying $\mat{U}$ by the singular values, a technique used to weight the latent dimensions considering the salience of the singular values,
did not result in any notable improvements in our experiments.

\subsection{Meta-Embedding Results}
\label{sec:res}

\begin{figure*}[t!]
\centering
\begin{subfigure}{.5\textwidth}
	\centering
	\includegraphics[height=50mm]{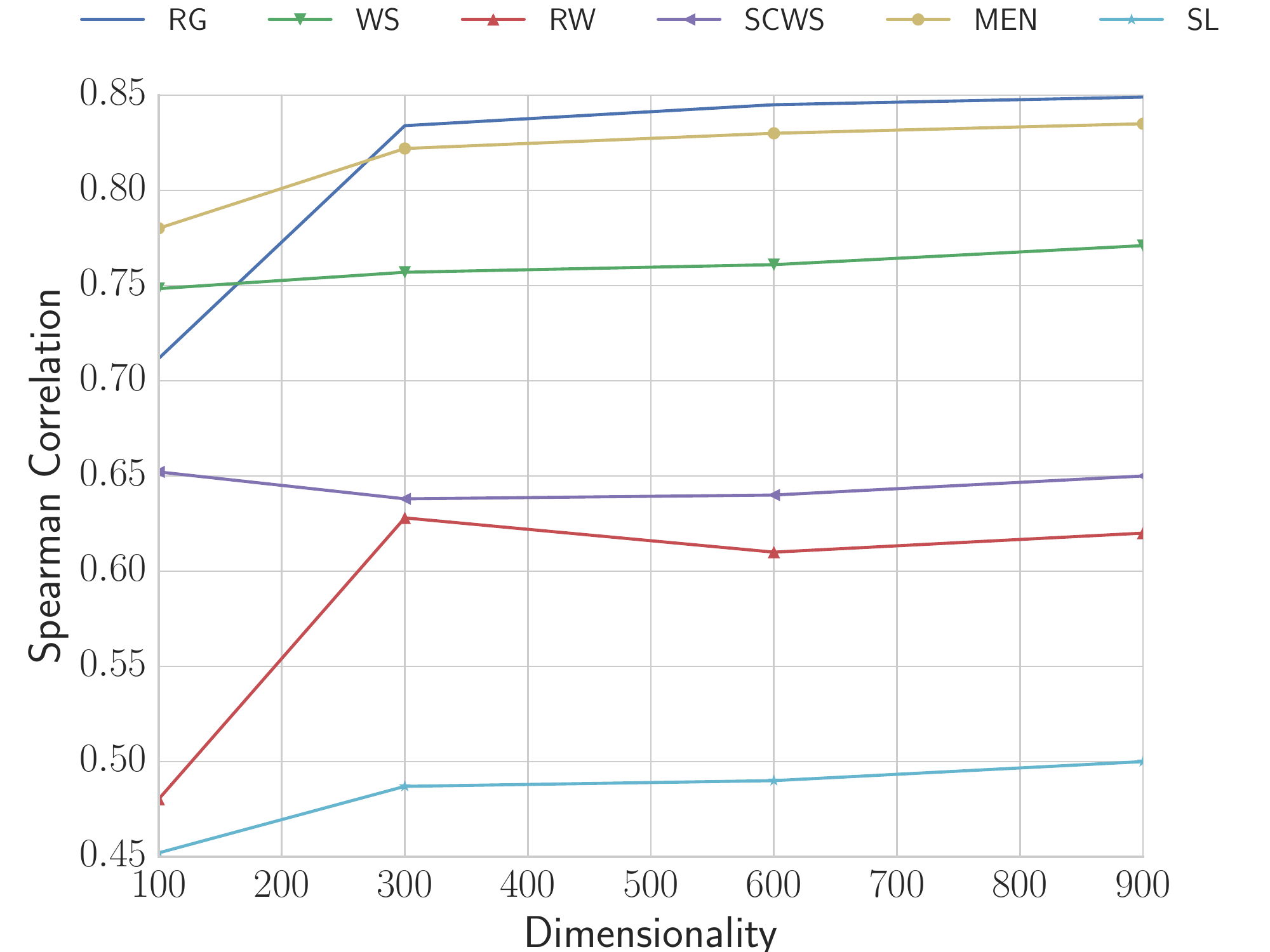}
	\caption{}
	\label{fig:k}
\end{subfigure}%
\begin{subfigure}{.5\textwidth}
	\centering
	\includegraphics[height=50mm]{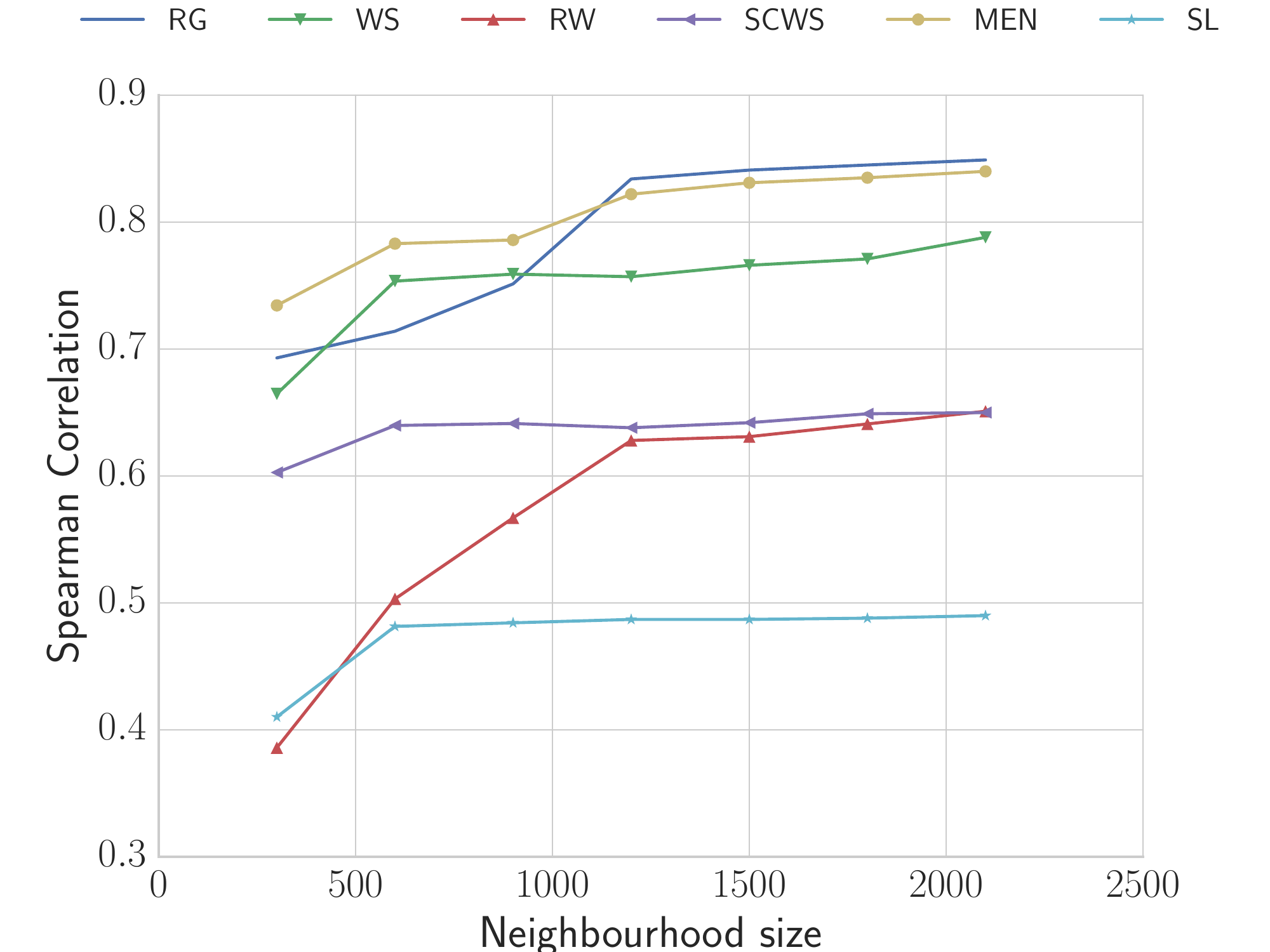}
	\caption{}
	\label{fig:n}
\end{subfigure}
\caption{Performance vs. dimensionality (neighbourhood size fixed at 1200) shown in (a), and vs. neighbourhood size (dimensionality fixed at 300) shown in (b) for meta embedding learning.}
\end{figure*}

\begin{table*}[t!]
\small
\centering
\begin{tabular}{p{1mm}p{2mm}|p{25mm}||p{4mm} p{4mm} p{5mm} p{5mm} p{5mm} p{5mm} p{4mm}|p{5mm} p{5mm}|p{5mm}|p{4mm}p{4mm}} 
 & & Model & RG & MC & WS & RW & SCWS & MEN & SL & GL & SE & DV & SA & MR\\ \hline
\parbox[t]{1mm}{\multirow{5}{*}{\rotatebox[origin=c]{90}{sources}}} 
& 1 &GloVe 	&  81.7 & 80.8 & 64.3 & 38.4 & 54.0 & 74.3 & 37.4 & 70.5 & 39.9 & 87.7 & 73.4 & 70.0 \\
& 2 & CBOW &  76.0 & 82.2 & 69.8 & 53.4 & 53.4 & 78.2 & 44.2 & 75.2 & 39.1 & 87.4 & 73.6 & 71.0 \\
& 3 & HLBL 	& 35.3 & 49.3 & 35.7 & 19.1 & 47.7 & 30.7 & 22.1 & 16.6 & 34.8 & 72.0 & 62.6 & 61.6 \\
& 4 & Huang 	& 51.3 & 58.8 & 58.0 & 36.4 & 63.7 & 56.1 & 21.7 & 8.3 & 35.2 & 76.0 & 64.8 & 60.9 \\
& 5 & CW 	& 29.9 & 34.3 & 28.4 & 15.3 & 39.8 & 25.7 & 15.6 & 4.7 & 34.6 & 75.6 & 62.7 & 61.4 \\ \cline{1-15}

\parbox[t]{2mm}{\multirow{15}{*}{\rotatebox[origin=c]{90}{ablation}}} 
& 6 & CONC (-GloVe) 	& 75.0 & 79.0 & 70.0 & 55.3 & 62.9 & 77.7 & 41.5 & 64.0 & 38.7 & 82.9 & 72.1 & 69.1 \\
& 7 & CONC (-CBOW) &  80.8 & 81.0 & 65.2 & 46.0 & 56.3 & 74.9 & 37.3 & 70.0 & 38.8 & 86.0 & 71.6 & 69.9 \\
& 8 & CONC (-HLBL) 	& 83.0 & 84.0 & 71.9 & 53.4 & 61.4 & $80.1^{*}$ & 41.6 & 72.7 & 39.5 & 84.9 & 71.0 & 69.4 \\
& 9 & CONC (-Huang) 	&  83.0 & 84.0 & 71.6 & 48.8 & 60.8 & $80.1^{*}$ & 41.9 & 72.8 & 40.0 & 86.7 & 71.2 & 69.1 \\
& 10 & CONC (-CW)  & 82.9 & 84.0 & 71.9 & 53.3 & 61.6 & $80.2^{*}$ & 41.6 & 72.6 & 39.6 & 84.9 & 72.3 & 69.9 \\  \cline{3-15}

& 11 & SVD (-GloVe) & 78.6 & 79.9 & 68.4 & 53.9 & 61.6 & 77.5 & 40.1 & 61.7 & 38.5 & 84.1 & 71.6 & 69.8 \\
& 12 & SVD (-CBOW) 	& 80.5 & 81.2 & 64.4 & 45.3 & 55.3 & 74.2 & 35.7 & 70.9 & 38.7 & 86.7 & 73.4 & 69.1 \\
& 13 & SVD (-HLBL) 	& 82.7 & 83.6 & 70.3 & 52.6 & 60.1 & $79.9^{*}$ & 39.6 & 73.5 & 39.8 & 87.3 & 73.2 & 70.4 \\
& 14 & SVD (-Huang) 	& 82.5 & 85.0 & 70.3 & 48.6 & 59.8 & $79.9^{*}$ & 39.9 & 73.7 & 40.0 & 87.3 & 73.5 & 70.8 \\
& 15 & SVD (-CW) 	& 82.5 & 83.9 & 70.4 & 52.5 & 60.1 & $80.0^{*}$ & 39.7 & 73.3 & 39.8 & 87.2 & 73.1 & 70.7 \\ \cline{3-15}

& 16 & Proposed (-GloVe) & 79.8 & 79.7 & 71.1 & 54.7 & 62.3 & 78.2 & 46.1 & $84.2^{*}$ & 39.8 & 85.4 & 72.2 & 70.2 \\
& 17 & Proposed (-CBOW) 	& 80.9 & 82.1 & 67.4 & $58.7^{*}$ & 58.7 & 75.7 & 45.2 & $85.2^{*}$ & 40.1 & 87.1 & 73.8 & 70.1 \\
& 18 & Proposed (-HLBL) 	& 82.1 & 86.1 & 71.3 & $58.3^{*}$ & 62.1 & $81.9^{*}$ & 34.8 & $86.3^{*}$ & 40.3 & 87.7 & 73.7 & 71.1 \\
& 19 & Proposed (-Huang) 	& 81.2 & 85.2 & 73.1 & 55.1 & 63.7 & $81.4^{*}$ & 42.3 & $82.6^{*}$ & 41.1 & 87.5 & 73.9 & 71.2 \\
& 20 & Proposed (-CW) 	& 83.1 & 84.8 & 72.5 & 58.5 & 62.3 & $81.1^{*}$ & 43.5 & $88.4^{*}$ & 41.9 & 87.8 & 71.6 & 71.1 \\ \cline{1-15}

\parbox[t]{2mm}{\multirow{5}{*}{\rotatebox[origin=c]{90}{ensemble}}}
& 21 &  CONC 		& 82.9 & 84.1 & 71.9 & 53.3 & 61.5 & $80.2^{*}$ & 41.6 & 72.9 & 39.6 & 84.9 & 72.4 & 69.9 \\ 
& 22 & SVD 			& 82.7 & 83.9 & 70.4 & 52.6 & 60.0 & $79.9^{*}$ & 39.7 & 73.4 & 39.7 & 87.2 & 73.4 & 70.7 \\
& 23 & 1\texttt{TO}N 	& 80.7 & 80.7 & 74.5 & $60.1^{*}$ & 61.6 & 73.5 & 46.4 & 76.8 & $42.3^{*}$ & 87.6 & 73.8 & 70.3 \\
& 24 & 1\texttt{TO}N+  & 82.7 & 85.0 & 75.3 & $61.6^{*}$ & 60.2 & 74.1 & 46.3 & 77.0 & 40.1 & 83.9 & 73.9 & 69.2 \\
& 25 & Proposed 		& \textbf{83.4} & \textbf{86.2} & $\mathbf{75.7}^{*}$ & $\mathbf{62.8}^{*}$ & \textbf{63.8} & $\mathbf{82.2}^{*}$ & \textbf{48.7} & $\mathbf{89.9}^{*}$ & $\mathbf{43.1}^{*}$ & $\mathbf{88.7}^{*}$ & \textbf{74.0} & \textbf{71.3} \\
\hline
\end{tabular}
\caption{Results on word similarity, analogy, relation and short-text classification tasks. For each task, the best performing method
is shown in bold. Statistically significant improvements over the best individual source embedding are indicated by an asterisk.}
\label{tbl:res}
\vspace{-4mm}
\end{table*}

Using the \textbf{MC} dataset, we find the best values for the neighbourhood size $n = 1200$ and dimensionality $d_{\cP} = 300$
for the \textbf{Proposed} method. We plan to publicly release our meta-embeddings on acceptance of the paper.

We summarise the experimental results for different methods on different tasks/datasets in Table~\ref{tbl:res}.
In Table~\ref{tbl:res}, rows 1-5 show the performance of the individual source embeddings. 
Next, we perform ablation tests (rows 6-20) where we hold-out one source embedding and use the other four with each meta-embedding method.
We evaluate statistical significance against best performing individual source embedding on each dataset.
For the semantic similarity benchmarks we use Fisher transformation to compute $p < 0.05$ confidence intervals for Spearman correlation coefficients.
In all other (classification) datasets, we used Clopper-Pearson binomial exact confidence intervals at $p < 0.05$. 

Among the individual source embeddings, we see that \textbf{GloVe} and \textbf{CBOW} stand out as the two best embeddings. 
This observation is further confirmed from ablation results, where the removal of \textbf{GloVe} or \textbf{CBOW} often results in a decrease in performance.
Performing SVD (rows 11-15) after concatenating, does not always result in an improvement.
SVD is a global projection that reduces the dimensionality of the meta-embeddings created via concatenation.
This result indicates that different source embeddings might require different levels of dimensionality reductions, and applying a single global projection
does not always guarantee improvements.
Ensemble methods that use all five source embeddings are shown in rows 21-25. 
\textbf{1\texttt{TO}N} and \textbf{1\texttt{TO}N+} are proposed by \newcite{Yin:ACL:2016}, and were detailed in Section~\ref{sec:related}. 
Because they did not evaluate on all tasks that we do here, to conduct a fair and consistent evaluation
we used their publicly available meta-embeddings\footnote{\url{http://cistern.cis.lmu.de/meta-emb/}} without retraining by ourselves.

Overall, from Table~\ref{tbl:res}, we see that the \textbf{Proposed} method (row 25) obtains the best performance in \emph{all} tasks/datasets.
In 6 out of 12 benchmarks, this improvement is statistically significant over the best single source embedding.
Moreover, in the \textbf{MEN} dataset (the largest among the semantic similarity benchmarks compared in Table~\ref{tbl:res} with 3000
word-pairs), and the \textbf{Google} dataset, the improvements of the \textbf{Proposed} method over the previously proposed \textbf{1\texttt{TO}N} and \textbf{1\texttt{TO}N+}
are statistically significant.

The ablation results for the \textbf{Proposed} method show that, although different source embeddings are important to different degrees, 
by using all source embeddings we can obtain the best results. 
Different source embeddings are trained from different resources and by optimising different objectives.
Therefore, for different words, the local neighbours predicted by different source embeddings will be complementary.
Unlike the other methods, the \textbf{Proposed} method never compares different source embeddings' vectors directly, but only via
the neighbourhood reconstruction weights. 
Consequently, the \textbf{Proposed} method is unaffected by relative weighting of source embeddings.
In contrast, the \textbf{CONC} is highly sensitive against the weighting. 
In fact, we confirmed that the performance scores of the CONC method were decreased by 3--10 points when we did not do the weight tuning described in Section~\ref{sec:eval-tasks}. 
The unnecessity of the weight tuning is thus a clear advantage of the Proposed method.

To investigate the effect of the dimensionality $d^{\cP}$ on the meta-embeddings learnt by the proposed method, in \autoref{fig:k},
we fix the neighbourhood size $N = 1200$ and measure the performance on semantic similarity measurement tasks when varying $d^{\cP}$.
Overall, we see that the performance peaks around $d^{\cP} = 300$. Such behaviour can be explained by the fact that
smaller $d^{\cP}$ dimensions are unable to preserve information contained in the
source embeddings, whereas increasing $d^{\cP}$ beyond the rank of the weight matrix $\mat{W}$ is likely to generate noisy eigenvectors.

In \autoref{fig:n}, we study the effect of increasing the neighbourhood size $n$ equally for all words in all source embeddings, while
fixing the dimensionality of the meta-embedding $d^{\cP} = 300$. Initially, performance increases with the neighbourhood size 
 and then saturates. This implies that in practice a small local neighbourhood is adequate to capture the differences in source embeddings. 
 
 \subsection{Complementarity of Resources}
 \label{sec:overlap}
 
 \begin{figure}[t]
\begin{center}
\includegraphics[width=80mm]{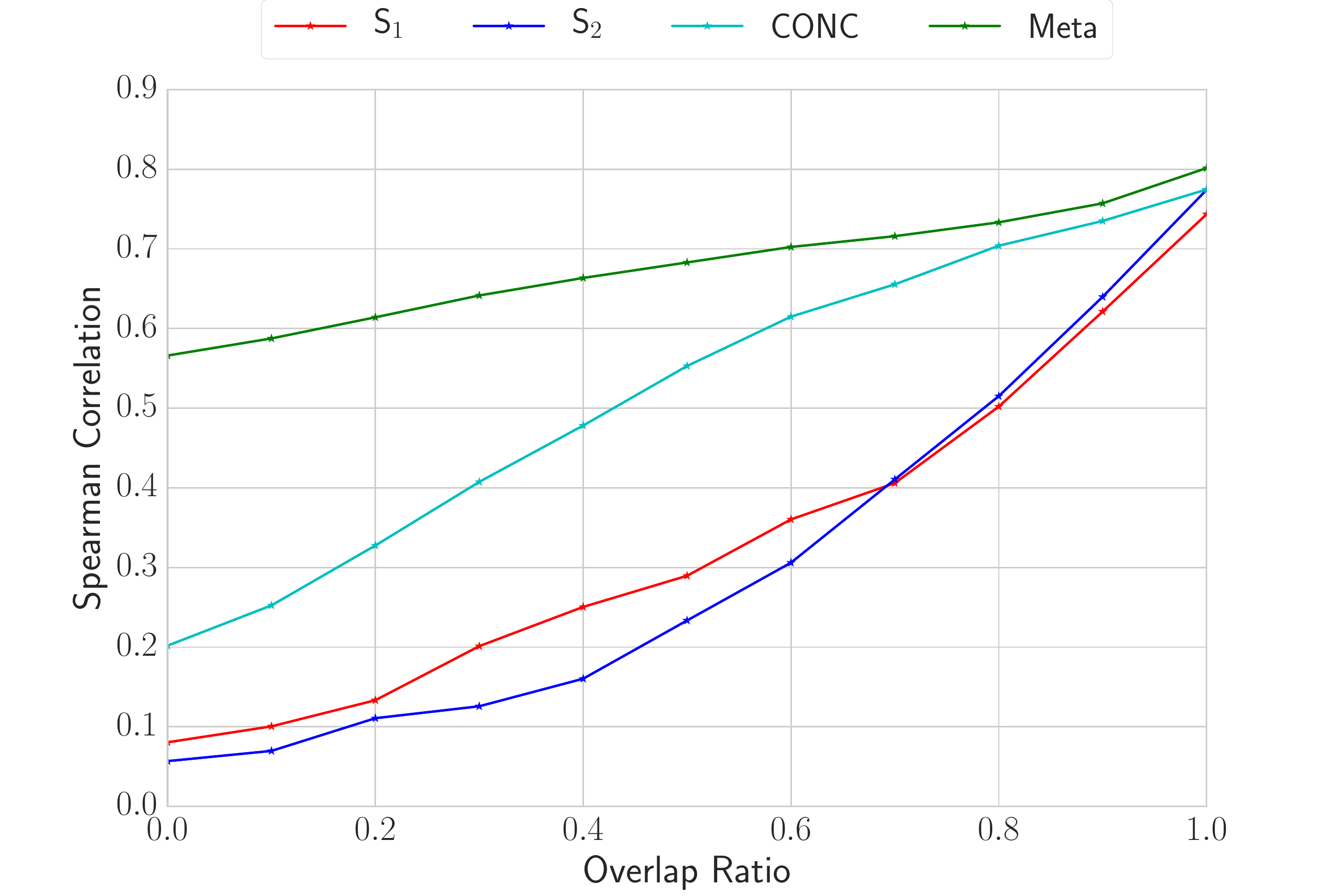}
\caption{Two word embedding learning algorithms trained on different but overlapping corpora to produce two source embeddings $S_{1}$ and $S_{2}$.
and their meta-embedding. 
}
\label{fig:overlap}
\end{center}
\vspace{-3mm}
\end{figure}

We have shown empirically in Section~\ref{sec:res} that using the proposed method it is possible to obtain superior meta-embeddings from a diverse set of source embeddings. One important scenario where meta-embedding could be potentially useful is when the source embeddings are trained on different complementary resources, where each resource share little common vocabulary. 
For example, one source embedding might have been trained on Wikipedia whereas a second source embedding might have been trained on tweets.

To evaluate the effectiveness of the proposed meta-embedding learning method under such settings, we design the following experiment.
We select \textbf{MEN} dataset, the largest among all semantic similarity benchmarks, which contains 751 unique words in 3000 human-rated word-pairs for semantic similarity. Next, we randomly split the set of words into two sets with different overlap ratios. We then select sentences from 2017 January dump of Wikipedia that contains words from only one of the two sets. We create two corpora of roughly equal number of sentences via this procedure for different overlap ratios. We train skip-gram with negative sampling (SGNS)~\cite{Milkov:2013} on one corpus to create source embedding $S_{1}$ and GloVe~\cite{Pennington:EMNLP:2014} on the other corpus to create source embedding $S_{2}$.
Finally, we use the proposed method to meta-embed $S_{1}$ and $S_{2}$.

Figure~\ref{fig:overlap} shows the Spearman correlation between the human similarity ratings and cosine similarities computed using the word embeddings on the \textbf{MEN} dataset for $S_{1}$, $S_{2}$ and their meta-embeddings created using the proposed method (Meta) and concatenation baseline (CONC).
From Figure~\ref{fig:overlap}, we see that the meta embeddings obtain the best performance across all overlap ratios.
The improvements are larger when the overlap between the corpora is smaller, and diminishes when the two corpora becomes identical.
This result shows that our proposed meta-embedding learning method captures the complementary information available in different source embeddings to create more accurate word embeddings. Moreover, it shows that by considering the local neighbourhoods in each of the source embeddings separately, we can obviate the need to predict embeddings for missing words in a particular source embedding, which was a limitation in the method proposed by \newcite{Yin:ACL:2016}.

 \nopagebreak
\section{Conclusion}
We proposed an unsupervised locally linear method for learning meta-embeddings from a given set of pre-trained source embeddings. 
Experiments on several NLP tasks show the accuracy of the proposed method, which outperforms previously proposed
meta-embedding learning methods on multiple benchmark datasets. In future, we plan to extend the proposed method to learn cross-lingual
meta-embeddings by incorporating both cross-lingual as well as monolingual information.

\bibliographystyle{aaai}
\bibliography{Embed}

\begin{thebibliography}{}

\bibitem[\protect\citeauthoryear{Bai}{2000}]{Bai:2000}
2000.
\newblock {\em Templates for the Solution of Algebraic Eigenvalue Problems: A
  Practical Guide}.
\newblock Society for the Industrial and Applied Mathematics.

\bibitem[\protect\citeauthoryear{Bansal, Gimpel, and
  Livescu}{2014}]{Bansal:ACL:2014}
Bansal, M.; Gimpel, K.; and Livescu, K.
\newblock 2014.
\newblock Tailoring continouos word representations for dependency parsing.
\newblock In {\em Proc. of ACL}.

\bibitem[\protect\citeauthoryear{Baroni, Dinu, and
  Kruszewski}{2014}]{baroni-dinu-kruszewski:2014:P14-1}
Baroni, M.; Dinu, G.; and Kruszewski, G.
\newblock 2014.
\newblock Don't count, predict! a systematic comparison of context-counting vs.
  context-predicting semantic vectors.
\newblock In {\em Proc. of ACL},  238--247.

\bibitem[\protect\citeauthoryear{Bruni \bgroup et al\mbox.\egroup }{2012}]{MEN}
Bruni, E.; Boleda, G.; Baroni, M.; and Tran, N.~K.
\newblock 2012.
\newblock Distributional semantics in technicolor.
\newblock In {\em Proc. of ACL},  136--145.

\bibitem[\protect\citeauthoryear{Chen \bgroup et al\mbox.\egroup
  }{2013}]{Chen:2013}
Chen, Y.; Perozzi, B.; Al-Rfou, R.; and Skiena, S.
\newblock 2013.
\newblock The expressive power of word embeddings.
\newblock In {\em Proc. of ICML Workshop}.

\bibitem[\protect\citeauthoryear{Collobert and
  Weston}{2008}]{Collobert:ICML:2008}
Collobert, R., and Weston, J.
\newblock 2008.
\newblock A unified architecture for natural language processing: Deep neural
  networks with multitask learning.
\newblock In {\em Proc. of ICML},  160 -- 167.

\bibitem[\protect\citeauthoryear{Duchi, Hazan, and
  Singer}{2011}]{Duchi:JMLR:2011}
Duchi, J.; Hazan, E.; and Singer, Y.
\newblock 2011.
\newblock Adaptive subgradient methods for online learning and stochastic
  optimization.
\newblock {\em Journal of Machine Learning Research} 12:2121--2159.

\bibitem[\protect\citeauthoryear{Goikoetxea, Agirre, and
  Soroa}{2016}]{AAAI:2016:Goikoetxea}
Goikoetxea, J.; Agirre, E.; and Soroa, A.
\newblock 2016.
\newblock Single or multiple? combining word representations independently
  learned from text and wordnet.
\newblock In {\em Proc. of AAAI},  2608--2614.

\bibitem[\protect\citeauthoryear{Halko, Martinsson, and
  Tropp}{2010}]{Halko:2010}
Halko, N.; Martinsson, P.~G.; and Tropp, J.~A.
\newblock 2010.
\newblock Finding structure with randomness: Probabilistic algorithms for
  constructung approximate matrix decompositions.
\newblock {\em SIAM REVIEW} 53(2):217 -- 288.

\bibitem[\protect\citeauthoryear{Hill \bgroup et al\mbox.\egroup
  }{2014}]{Hill:NIPS:2014}
Hill, F.; Cho, K.; Jean, S.; Devin, C.; and Bengio, Y.
\newblock 2014.
\newblock Not all neural embeddings are born equal.
\newblock In {\em NIPS workshop}.

\bibitem[\protect\citeauthoryear{Hill \bgroup et al\mbox.\egroup
  }{2015}]{Hill:ICLR:2015}
Hill, F.; Cho, K.; Jean, S.; Devin, C.; and Bengio, Y.
\newblock 2015.
\newblock Embedding word similarity with neural machine translation.
\newblock In {\em ICLR Workshop}.

\bibitem[\protect\citeauthoryear{Hill, Reichart, and Korhonen}{2015}]{SimLex}
Hill, F.; Reichart, R.; and Korhonen, A.
\newblock 2015.
\newblock Simlex-999: Evaluating semantic models with (genuine) similarity
  estimation.
\newblock {\em Computational Linguistics} 41(4):665--695.

\bibitem[\protect\citeauthoryear{Huang \bgroup et al\mbox.\egroup
  }{2012}]{Huang:ACL:2012}
Huang, E.~H.; Socher, R.; Manning, C.~D.; and Ng, A.~Y.
\newblock 2012.
\newblock Improving word representations via global context and multiple word
  prototypes.
\newblock In {\em Proc. of ACL},  873--882.

\bibitem[\protect\citeauthoryear{Jurgens \bgroup et al\mbox.\egroup
  }{2012}]{SemEavl2012:Task2}
Jurgens, D.~A.; Mohammad, S.; Turney, P.~D.; and Holyoak, K.~J.
\newblock 2012.
\newblock Measuring degrees of relational similarity.
\newblock In {\em Proc. of SemEval}.

\bibitem[\protect\citeauthoryear{Kibriya and Frank}{2007}]{BallTree}
Kibriya, A.~M., and Frank, E.
\newblock 2007.
\newblock An empirical comparison of exact nearest neighbout algorithms.
\newblock In {\em Proc. of PKDD},  140--151.

\bibitem[\protect\citeauthoryear{Luo \bgroup et al\mbox.\egroup
  }{2014}]{Luo:AAAI:2014}
Luo, Y.; Tang, J.; Yan, J.; Xu, C.; and Chen, Z.
\newblock 2014.
\newblock Pre-trained multi-view word embedding using two-side neural network.
\newblock In {\em Proc. of AAAI},  1982--1988.

\bibitem[\protect\citeauthoryear{Luong, Socher, and
  Manning}{2013}]{Luong:CoNLL:2013}
Luong, M.-T.; Socher, R.; and Manning, C.~D.
\newblock 2013.
\newblock Better word representations with recursive neural networks for
  morphology.
\newblock In {\em CoNLL}.

\bibitem[\protect\citeauthoryear{Mikolov \bgroup et al\mbox.\egroup
  }{2013}]{Mikolov:NIPS:2013}
Mikolov, T.; Sutskever, I.; Chen, K.; Corrado, G.~S.; and Dean, J.
\newblock 2013.
\newblock Distributed representations of words and phrases and their
  compositionality.
\newblock In {\em Proc. of NIPS},  3111--3119.

\bibitem[\protect\citeauthoryear{Mikolov, Chen, and Dean}{2013}]{Milkov:2013}
Mikolov, T.; Chen, K.; and Dean, J.
\newblock 2013.
\newblock Efficient estimation of word representation in vector space.
\newblock In {\em Proc. of ICLR}.

\bibitem[\protect\citeauthoryear{Miller and Charles}{1998}]{MC}
Miller, G., and Charles, W.
\newblock 1998.
\newblock Contextual correlates of semantic similarity.
\newblock {\em Language and Cognitive Processes} 6(1):1--28.

\bibitem[\protect\citeauthoryear{Mnih and Hinton}{2009}]{Mnih:HLBL:NIPS:2008}
Mnih, A., and Hinton, G.~E.
\newblock 2009.
\newblock A scalable hierarchical distributed language model.
\newblock In Koller, D.; Schuurmans, D.; Bengio, Y.; and Bottou, L., eds., {\em
  Proc. of NIPS}.
\newblock  1081--1088.

\bibitem[\protect\citeauthoryear{Pang and Lee}{2005}]{Pang:ACL:2005}
Pang, B., and Lee, L.
\newblock 2005.
\newblock Seeing stars: Exploiting class relationships for sentiment
  categorization with respect to rating scales.
\newblock In {\em Proc. of ACL},  115--124.

\bibitem[\protect\citeauthoryear{Pennington, Socher, and
  Manning}{2014}]{Pennington:EMNLP:2014}
Pennington, J.; Socher, R.; and Manning, C.~D.
\newblock 2014.
\newblock Glove: global vectors for word representation.
\newblock In {\em Proc. of EMNLP},  1532--1543.

\bibitem[\protect\citeauthoryear{Rubenstein and Goodenough}{1965}]{RG}
Rubenstein, H., and Goodenough, J.
\newblock 1965.
\newblock Contextual correlates of synonymy.
\newblock {\em Communications of the ACM} 8:627--633.

\bibitem[\protect\citeauthoryear{Socher \bgroup et al\mbox.\egroup
  }{2013}]{socher-EtAl:2013:EMNLP}
Socher, R.; Perelygin, A.; Wu, J.; Chuang, J.; Manning, C.~D.; Ng, A.; and
  Potts, C.
\newblock 2013.
\newblock Recursive deep models for semantic compositionality over a sentiment
  treebank.
\newblock In {\em Proc. of EMNLP},  1631--1642.

\bibitem[\protect\citeauthoryear{Tian \bgroup et al\mbox.\egroup
  }{2016}]{Tian:arxiv:2016}
Tian, Y.; Kullkarni, V.; Perozzi, B.; and Skiena, S.
\newblock 2016.
\newblock On the convergent properties of word embedding methods.
\newblock {\em arXiv}.

\bibitem[\protect\citeauthoryear{Tsuboi}{2014}]{tsuboi:2014:EMNLP2014}
Tsuboi, Y.
\newblock 2014.
\newblock Neural networks leverage corpus-wide information for part-of-speech
  tagging.
\newblock In {\em Proc. of EMNLP},  938--950.

\bibitem[\protect\citeauthoryear{Turian, Ratinov, and
  Bengio}{2010}]{Turian:ACL:2010}
Turian, J.; Ratinov, L.; and Bengio, Y.
\newblock 2010.
\newblock Word representations: A simple and general method for semi-supervised
  learning.
\newblock In {\em Proc. of ACL},  384 -- 394.

\bibitem[\protect\citeauthoryear{Turney and Pantel}{2010}]{Turney:JAIR:2010}
Turney, P.~D., and Pantel, P.
\newblock 2010.
\newblock From frequency to meaning: Vector space models of semantics.
\newblock {\em Journal of Aritificial Intelligence Research} 37:141 -- 188.

\bibitem[\protect\citeauthoryear{Vylomova \bgroup et al\mbox.\egroup
  }{2016}]{Vylomova:ACL:2016}
Vylomova, E.; Rimell, L.; Cohn, T.; and Baldwin, T.
\newblock 2016.
\newblock Take and took, gaggle and goose, book and read: Evaluating the
  utility of vector differences for lexical relational learning.
\newblock In {\em ACL},  1671--1682.

\bibitem[\protect\citeauthoryear{Yin and Sch\"{u}tze}{2016}]{Yin:ACL:2016}
Yin, W., and Sch\"{u}tze, H.
\newblock 2016.
\newblock Learning meta-embeddings by using ensembles of embedding sets.
\newblock In {\em Proc. of ACL},  1351--1360.

\bibitem[\protect\citeauthoryear{Zou \bgroup et al\mbox.\egroup
  }{2013}]{Zou:EMNLP:2013}
Zou, W.~Y.; Socher, R.; Cer, D.; and Manning, C.~D.
\newblock 2013.
\newblock Bilingual word embeddings for phrase-based machine translation.
\newblock In {\em Proc. of EMNLP},  1393--1398.

\end{thebibliography}

\end{document}